%% file: sample-sigconf.tex
\documentclass[sigconf]{acmart}

\AtBeginDocument{%
  \providecommand\BibTeX{{%
    \normalfont B\kern-0.5em{\scshape i\kern-0.25em b}\kern-0.8em\TeX}}}

\setcopyright{acmcopyright}
\copyrightyear{2020}
\acmYear{2020}
\usepackage{latexsym}

\usepackage{soul}
\usepackage{url}
\usepackage{hyperref}
\usepackage[utf8]{inputenc}
\usepackage{caption}
\usepackage{graphicx}
\usepackage{amsmath}
\usepackage{booktabs}
\usepackage{tabu}

\usepackage{enumitem}
\usepackage{siunitx}
\usepackage{multirow}
\usepackage{algorithm}
\usepackage{setspace}
\usepackage{algcompatible}
\usepackage[noend]{algpseudocode}
\makeatletter
\def\BState{\State\hskip-\ALG@thistlm}
\makeatother

\algdef{SE}[SUBALG]{Indent}{EndIndent}{}{\algorithmicend\ }%
\algtext*{Indent}
\algtext*{EndIndent}

\copyrightyear{2021}
\acmYear{2021}
\setcopyright{acmlicensed}
\acmConference[WSDM '21] {Proceedings of the Fourteenth ACM International Conference on Web Search and Data Mining}{March 8--12, 2021}{Virtual Event, Israel}
\acmBooktitle{Proceedings of the Fourteenth ACM International Conference on Web Search and Data Mining (WSDM '21), March 8--12, 2021, Virtual Event, Israel}
\acmPrice{15.00}
\acmDOI{10.1145/3437963.3441814}
\acmISBN{978-1-4503-8297-7/21/03}
\newcommand\METHOD{Self-Pretraining~}
\newcommand\ADRDT{ADR~}
\newcommand\EARTHQUAKEDT{Earthquake~}
\newcommand\PRODUCTDT{Product~}

\setcopyright{none}
\begin{document}
\fancyhead{}
\title{Semi-Supervised Text Classification via Self-Pretraining}


\author{Payam Karisani}
\affiliation{Emory University}
\email{payam.karisani@emory.edu}

\author{Negin Karisani}
\affiliation{Purdue University}
\email{nkarisan@purdue.edu}


\begin{abstract}
    We present a neural semi-supervised learning model termed Self-Pretraining. Our model is inspired by the classic self-training algorithm. However, as opposed to self-training, Self-Pretraining is threshold-free, it can potentially update its belief about previously labeled documents, and can cope with the semantic drift problem. Self-Pretraining is iterative and consists of two classifiers. In each iteration, one classifier draws a random set of unlabeled documents and labels them. This set is used to initialize the second classifier, to be further trained by the set of labeled documents. The algorithm proceeds to the next iteration and the classifiers' roles are reversed. To improve the flow of information across the iterations and also to cope with the semantic drift problem, Self-Pretraining employs an iterative distillation process, transfers hypotheses across the iterations, utilizes a two-stage training model, uses an efficient learning rate schedule, and employs a pseudo-label transformation heuristic. We have evaluated our model in three publicly available social media datasets. Our experiments show that Self-Pretraining outperforms the existing state-of-the-art semi-supervised classifiers across multiple settings. Our code is available at \url{https://github.com/p-karisani/self_pretraining}.
\end{abstract}




% %% A "teaser" image appears between the author and affiliation
% %% information and the body of the document, and typically spans the

\maketitle

\input{doc-body.tex}

\bibliographystyle{ACM-Reference-Format}
\bibliography{sample-base}

\end{document}

%% file: doc-body.tex
\section{Introduction} \label{sec:intro}
\input{doc-intro.tex}

\vspace{-0.1cm}
\section{Related Work} \label{sec:rel-work}
\input{doc-related-work.tex}

\vspace{-0.1cm}
\section{Semi-Supervised Learning via \METHOD} \label{sec:method}
\input{doc-method.tex}

\section{Experimental Setup} \label{sec:setup}
\input{doc-experiment-setup.tex}

\vspace{-0.1cm}
\section{Results and Discussion} \label{sec:result}
\input{doc-results.tex}

\section{Conclusions} \label{sec:conclusion}
\input{doc-conclusion.tex}

% \section*{Acknowledgments} \label{sec:ack}
\input{doc-ack.tex}

%% file: doc-intro.tex
Semi-supervised text classifiers have achieved remarkable success in the past few years due to the high capacity of neural networks in generalization. Even though modern classifiers usually rely on large training sets, the introduction of contextual word embeddings and language model pretraining \cite{elmo,bert,T5} has tremendously reduced the need for manual data annotation. However, the state-of-the-art neural models are still prone to overfitting, particularly in the areas with sparse and specialized language models. These areas include, but are not limited to: legal domain \cite{legal-domain}, medical domain \cite{biobert}, and social media domain \cite{social-media}.

Depending on the task at hand, one solution to address this issue is to automatically construct a large--and perhaps noisy--dataset \cite{sarcasm}, however, this is not always feasible \cite{crisis-wit}. A more methodical approach is to employ the techniques that improve generalization. These techniques include exploiting neural word embeddings \cite{wespad}, data augmentation \cite{sarcasm-augment}, and domain adaptation \cite{crisis-domain-adapt}. Exploiting unlabeled data \cite{self-train,co-train} is also a complementary approach. In this study, we add to the body of literature on semi-supervised learning by employing the properties of neural networks and proposing a novel way to utilize unlabeled data. We focus on one of the areas that reportedly suffers from the lack of enough training data, i.e., the social media mining. In addition to the lack of training data, the progress in this domain is further hindered by the short document lengths, informal language model, and typically ambiguous choice of vocabularies. These qualities make the social media tasks a suitable test bed for evaluating semi-supervised learning algorithms.

Our algorithm, termed \METHOD\!, is inspired by the self-training paradigm \cite{self-train}. Similar to self-training, our algorithm is iterative and in each iteration selects a set of unlabeled documents to label. However, as opposed to self-training, our algorithm is threshold-free. Thus, it does not rank the unlabeled documents based on their prediction confidences. This makes our algorithm particularly suitable for the neural network models due to their poorly calibrated outputs \cite{neural-conf-2}. Additionally, our algorithm is able to cope with the semantic drift problem \cite{nell}. That is, it is resilient to the noise in the \textit{pseudo-labels} as the number of iterations increases and the error rate of the underlying classifier rises. Furthermore, \METHOD is able to potentially revise the labels of the previously labeled documents. To achieve these, our model employs an iterative distillation process, i.e., in each iteration, the information obtained in the previous iterations is distilled into the classifier. It transfers a hypothesis across iterations, and utilizes a two-stage learning model, where the set of pseudo-labels is used to initialize the classifier, and the set of labeled documents is used to finetune the classifier. Additionally, \METHOD adapts a novel learning rate schedule to efficiently integrate the two sets of noisy and noise-free training examples. Finally, in order to further mitigate the impact of noisy pseudo-labels in every iteration, our model transforms the distribution of pseudo-labels such that it reflects the distribution of the labels in the previous iterations.

Our experiments in three publicly available Twitter datasets show that \METHOD outperforms the state of the art in multiple settings where only a few hundred labeled documents are available. This is significant, considering that the underlying classifier of our algorithm and all the baseline models is BERT \cite{bert} which already uses the language model pretraining, and therefore, makes any improvement over the baselines very challenging. We also carry out a comprehensive set of experiments to better understand the qualities of \METHOD\!. Particularly, we demonstrate the robustness of our model against the noise in the pseudo-labels. The contributions of our study are as follows:
    \textbf{1)} We propose a novel semi-supervised learning framework termed \METHOD\!. Our model is based on the self-training paradigm, however, it is threshold-free, it can cope with the semantic drift problem, and can also revise the previously labeled documents. To our knowledge, \METHOD is the first model that addresses these drawbacks in a unified framework.
    \textbf{2)} We propose a novel learning rate schedule to effectively integrate the optimization procedure with our two-stage semi-supervised learning process.
    \textbf{3)} In order to further mitigate the semantic drift problem, we model the class distribution of the pseudo-labels as a stochastic process across the bootstrapping iterations, and propose a novel approach to transform the class distributions.
    \textbf{4)} We carry out a comprehensive set of experiments across three publicly available Twitter datasets, and demonstrate that our model outperforms several state-of-the-art baselines in multiple settings.
    
Our research clearly pushes the state of the art in semi-supervised text classification. We believe the ideas presented in our paper can be applied to other domains, e.g., image classification. Future work may explore this direction. In the next section, we provide an overview of the related studies and highlight the qualities of \METHOD\!.

%% file: doc-related-work.tex
\noindent\textbf{Unlabeled data in semi-supervised learning.} Unlabeled data can be exploited in multiple ways. It can be used as a meta-source of information \cite{var-pretrain}, it can be used as a regularizer \cite{UDA}, or it can be used in a domain adaptation setting to correlate the source and target data \cite{coral}. A more recent interest in literature is \textit{self-supervision}, where a self-contained task is defined such that no manual annotation is required. Instances of such tasks are language model pretraining \cite{elmo,bert} in NLP, and contrastive learning in image processing \cite{consistency-loss,consistency}. From a different perspective, self-supervision studies can be categorized into task-agnostic \cite{lm-pretrain-fewshot} and task-specific \cite{do-pretrain} approaches. This has given rise to the notion of ``pretrain, then finetune'' the model. We integrate this paradigm into the self-training algorithm.

\noindent\textbf{Bootstrapping in semi-supervised learning.} Self-training is the oldest approach to semi-supervised learning \cite{semi-supervised-book} dating back to 1965 \cite{self-train-0}. This idea re-emerged in the seminal work of Yarowsky \cite{self-train} for NLP tasks in 1995, and also once more in the computer vision community in 2013 as \textit{pseudo-labeling} \cite{pseudo-label}. This algorithm is a wrapper that repeatedly uses a supervised algorithm as the underlying model. There are multiple assumptions under which self-training--and in general semi-supervised learning--is expected to perform well. For instance, the \textit{smoothness assumption} that states if the two data points $x_1$ and $x_2$ are close, then their predictions $y_1$ and $y_2$ should be also close--this assumption has been the basis of algorithms such as MixUp \cite{mixup} and MixMatch \cite{mixmatch}. As we discuss in the next section, one unsatisfactory aspect of self-training is that it relies on the properties of the underlying predictive model, e.g., the model output distributions. There have been attempts to address this drawback. For instance, throttling \cite{semiupervised-nlp-book} can be used to dampen the effect of noisy candidates, or in the context of transductive learning, the density of the unlabeled data points can be incorporated to mitigate this issue \cite{neural-pseudo-label-2}.

In the past few years, studies have explored the efficacy of the neural networks as the underlying predictive model in self-training. A neural network variant of co-training \cite{co-train} is proposed in \cite{deep-co-train}. In \cite{co-decomp}, the authors propose a framework to integrate human knowledge with co-training. In \cite{reinforced-co-train}, a reinforcement learning variant of co-training is proposed. In \cite{neural-self-train-baselines}, a neural network variant of tri-training with disagreement \cite{tri-train-d} is presented, and it is shown that the combination is a surprisingly strong baseline in the domain adaptation setting. The authors in \cite{curriculum-selftrain} propose to use percentile scores instead of the confidence scores to select the best pseudo-labels; and the authors in \cite{uncertain-selftrain} employ Bayesian neural networks to select the most and the least confident pseudo-labels in every iteration. In \cite{self-train-queue}, a new document sampling strategy for self-training is proposed. The model, in addition to the classifier confidence, employs the training epochs in which the unlabeled documents are approximately correctly labeled. In \cite{deep-pseudo-label}, the authors propose to integrate MixUp \cite{mixup} with the oversampling of the labeled training examples. They show that self-training is indeed a very strong baseline comparing to the common regularization and data augmentation techniques. In comparison to these studies, \METHOD is the first model that employs model distillation \cite{distill-2} along a hypothesis to transfer information across iterations, enabling it to potentially revise the pseudo-labels. It integrates the pretraining/finetuning paradigm with self-training, utilizes an efficient optimization procedure along a perturbation technique to mitigate the negative impact of noisy pseudo-labels.

\noindent\textbf{Other closely related studies.} In addition to the studies above, \METHOD is also related to the studies on model distillation \cite{distill-2} and temporal ensembling \cite{temporal-ensemble}. Model distillation was proposed in \cite{distill-1,distill-2} to transfer the knowledge from one model to another model. In \cite{big-distill}, the authors show that transferring the knowledge of a big network, trained by a self-supervised task, to a small network improves generalization. Their main contribution is to show that big models are trained easier, and therefore, can be used as a proxy to train small networks. Their model is not iterative, and does not explore the unlabeled data to extract new information. Born-again networks were proposed in \cite{born-again}, the authors show that simply distilling a neural network into itself improves performance. Their model is not a semi-supervised algorithm, and is not proposed to exploit unlabeled data. The authors in \cite{noisy-student} show that the regular neural self-training algorithm can be improved by adding noise to the model. Similar to our work, they allow the pseudo-labels evolve over iterations. Beyond this step, they don't propose any modification to the self-training algorithm. Additionally, the efficacy of their model is not explored in the semi-supervised setting. A very close approach to this study is presented in \cite{noisy-selftrain}, where the authors again show that adding noise to the inner representation of the model enhances the self-training performance. Temporal ensembling was proposed in \cite{temporal-ensemble}. The authors propose to maintain the per-sample prediction average of the unlabeled data across the epochs and constrain the prediction variance. Their model is not based on self-training, has no strategy to separate labeled from unlabeled data, and becomes unwieldy when using large datasets. The authors in \cite{mean-teacher} resolve the high complexity of temporal ensembling by updating the weights of the model across the epochs, instead of storing the predictions.

%% file: doc-method.tex
We begin this section by providing an overview of \METHOD\! and highlighting its differences from the self-training algorithm. Then we introduce a series of strategies to overcome the drawbacks of the vanilla \METHOD\!\footnote{We focus on the binary classification problems.}.

In the self-training algorithm \cite{self-train}, a small set of labeled documents $L$ and a large set of unlabeled documents $U$ are available for training. The algorithm is iterative and in each iteration the predictive model $M$ is trained on the current set $L$, and is used to probabilistically label the current set $U$. Given the hyper-parameter $\theta$ as the minimum confidence threshold, the most confidently labeled documents in $U$ and their associated \textit{pseudo-labels} are selected to be augmented with the set $L$. This procedure is repeated till a certain criterion is met. There are three drawbacks with this algorithm: 1) The semantic drift problem \cite{nell}, where the increasingly negative impact of noisy pseudo-labels overshadows the benefit of incorporating unlabeled data. 2) Reliance on the model calibration. If the underlying classifier is unable to accurately model the class distributions, then, it will fail to properly rank the candidate documents, e.g., in the case of neural networks \cite{neural-conf-2}. 3) Being unable to revise the pseudo-labels once they are assigned to the unlabeled documents and augmented with the set of labeled documents. Even though there exist techniques to address these challenges under certain conditions, e.g., throttling \cite{semiupervised-nlp-book} for the poor model calibration or mutual exclusive bootstrapping \cite{semantic-drift} for the semantic drift, to our knowledge, \METHOD is the first unified framework to address all three.

Our algorithm is iterative and utilizes two neural networks as the underlying classifiers. Algorithm \ref{alg:summary} illustrates \METHOD in its basic form. Initially, the set $L$ is used to train the network $M_1$ (Line \ref{alg-line:init-train}), then the parameters of $M_1$ are copied to the network $M_2$ (Line \ref{alg-line:copy}). In the next step, a set of unlabeled documents are randomly drawn from $U$ (Line \ref{alg-line:sample}). This set is labeled by $M_2$ and used along the set $L$ to retrain\footnote{Note that by definition, the neural self-training requires reinitialization and retraining in every iteration \cite{neural-self-train-baselines}, thus our algorithm is comparable to other self-training models in terms of runtime.} $M_1$ (Line \ref{alg-line:retrain}). The role of the two networks is reversed in the next iteration. In each iteration, the sample size is increased by $k$ (Line \ref{alg-line:sample-inc}), and the algorithm stops when the sample set covers the entire set $U$.  Finally, the ensemble of $M_1$ and $M_2$ can be used to label the unseen documents--we used the mean of their class predictions.
\begin{algorithm}
{\small
\caption{Overview of Vanilla \METHOD}\label{alg:summary}
\begin{algorithmic}[1]
\algrenewcommand\algorithmicindent{7pt}
\Function{self\_pretraining}{$L, U, k$}
    \State $M_1 \gets train\_model(L)$ \label{alg-line:init-train}
    \State $sample\_size \gets 0$
    \Repeat
        \State $M_2 \gets copy\_model(M_1)$ \label{alg-line:copy}
        \State $sample\_size \gets sample\_size + k$ \label{alg-line:sample-inc}
        \State $C \gets random\_sample(U, sample\_size)$ \label{alg-line:sample}
        \State $M_1 \gets train\_model(\{(C,M_2(C)) \cup L\})$ \label{alg-line:retrain}
    \Until {$sample\_size < \left|U\right|$}
    \State \textbf{return} $M_1, M_2$
\EndFunction
\end{algorithmic}}
\end{algorithm}

Algorithm \ref{alg:summary} has two advantages: 1) To select the pseudo-labels the class distribution is not taken into account, therefore, there is no constraint on the classifier capacity in ranking the unlabeled documents. Additionally, this prevents the model from repeatedly selecting a fixed set of unlabeled documents in every iteration--i.e., the set of highly confident pseudo-labels. 2) The information that is transferred across the iterations is in the form of a hypothesis rather than a set of fixed pseudo-labels. Therefore, the model belief about the pseudo-labels can evolve over time--the pseudo-labels are not augmented with the set of labeled documents. On the other hand, this algorithm has one substantial disadvantage, and that is the problem of semantic drift. In fact, randomly sampling from the set of unlabeled documents exacerbates this problem by introducing noisy labels and pushing the transferred hypothesis towards a sub-optimal point. In the following, we exploit the neural network properties and introduce a series of strategies to cope with this problem and also to enhance the flow of information across the iterations.

\subsection{Hypothesis Transfer and Iterative Distillation} \label{sub-sec:method-distill}
\METHOD transfers a hypothesis--a learned function--from one iteration to the next iteration. In each iteration, this hypothesis is used to form a new one by creating a set of pseudo-labels and augmenting them with the set of labeled documents. Even though the ultimate criterion is maximizing the model utility, the short term goal in each iteration is not necessarily making accurate predictions but to carefully transfer the knowledge from one model to another. These two processes are not necessarily in accordance with each other, since the former may rely on the learner outcome and the latter may rely on the learning procedure itself. Thus, the classifier labels, even though informative, are not expressive enough to transfer the entire knowledge from one iteration to the next one.

The authors in \cite{distill-1,distill-2} propose an algorithm called model distillation to transfer the knowledge from a large model (called \textit{teacher}) to a small model (called \textit{student}). Model distillation is based on the argument that the class distribution carries a significant amount of information regarding the classifier decision boundary. For instance, given a document $d$ that is labeled positive, it is nontrivial information to know that if the class prediction was 95\% positive or 65\% positive. The authors in \cite{distill-2} use model distillation to transfer knowledge from one network to another by modifying the softmax layer as follows:
\begin{equation} \label{eq:softmax}
\setlength{\jot}{0pt}
\setlength{\abovedisplayskip}{5pt}
\setlength{\belowdisplayskip}{5pt}
\medmuskip=0mu
\thinmuskip=0mu
\thickmuskip=0mu
\nulldelimiterspace=0pt
\scriptspace=0pt
a_i=\frac{\exp \frac{z_i}{T}}{\sum_{j}^{}\exp \frac{z_j}{T}}
\end{equation}
where $z_i$ is the last layer \textit{i-th} logit, $j$ is the number of classes, and $a_i$ is the class prediction. The hyper-parameter $T$ is called temperature and is introduced to smooth the class predictions. A higher temperature results in a higher entropy in predictions. This is particularly desirable, since neural networks are known to have a low entropy in their predictions \cite{neural-conf-2}. 

Given the argument above, we employ model distillation in \METHOD\!, and effectively distill the previous iterations into the student network $M_1$. Thus, in each iteration, instead of using the teacher--$M_2$--hard predictions on unlabeled documents, we use the soft predictions along the set $L$ to train the student network--Algorithm \ref{alg:summary}, Line \ref{alg-line:retrain}.

\subsection{Two-Stage Semi-Supervised Learning} \label{sub-sec:method-pretrain}
As we mentioned earlier, self-training suffers from the semantic drift problem. This problem occurs when the errors primarily caused by the pseudo-labels accumulate across iterations and ultimately distort the classifier boundary. Even though the minimum confidence threshold $\theta$ can potentially prevent spurious pseudo-labels from entering the training set, as the set $L$ grows in size the probability of mislabeling documents increases correspondingly. This problem is even severer in our model, since it is threshold-free. One naive solution is to assign a lower weight to the pseudo-labels, however, we observed in our experiments that this approach is not effective enough to resolve the underlying problem.

To mitigate this problem, one solution is to process the set of pseudo-labels and decouple the information that contradicts the information stored in the set $L$. Erasing this section of the pseudo-labels can lower the error rate and subsequently improve the hypothesis in the current iteration. To accomplish this, we exploit the catastrophic forgetting phenomenon in the neural networks \cite{catastrophic-1,catastrophic-2}. Catastrophic forgetting occurs in the continual learning setting where a network is trained on a series of tasks. Each training procedure updates the parameters of the model to meet the requirements of the objective function, and the updates in the current task may contradict and erase the information related to the previous tasks. This effect is typically undesirable, however, in the context of \METHOD\!, we use this mechanism as a proxy to build a hierarchy of information in the network. Therefore, we make a small modification in Algorithm \ref{alg:summary}. Instead of aggregating the set of pseudo-labels with the set of labeled documents--Line \ref{alg-line:retrain}--we first use the set of pseudo-labels to initialize--train--the current network $M_1$, and then further train it using the set of labeled documents.

Decomposing the training procedure into two stages introduces a new challenge, and that is the possibility of completely updating the network parameters in order to learn the regularities in the set of labeled documents. To avoid this, we propose to use the following objective function while training the model $M_1$ using the set $L$:
\begin{equation} \label{eq:loss}
\setlength{\jot}{0pt}
\setlength{\abovedisplayskip}{0pt}
\setlength{\belowdisplayskip}{0pt}
\medmuskip=0mu
\thinmuskip=0mu
\thickmuskip=0mu
\nulldelimiterspace=0pt
\scriptspace=0pt
\begin{split}
\mathcal{L}= &(1-\lambda)(-\sum_{i=1}^{N}[y_{i}\log a_{i}+(1-y_{i})\log (1-a_{i})]) + \\ 
&\lambda(-\sum_{i=1}^{N}[q_{i}\log a'_{i}+(1-q_{i})\log (1-a'_{i})])
\end{split}
\end{equation}
where $N$ is the number of the documents in the set $L$, $y_{i}$ is the true label of the document $d_{i}$, $a_{i}$ is the class prediction of $M_1$ for $d_{i}$, $a'_{i}$ is the class prediction of $M_1$ for $d_{i}$ with a high temperature as described in Section \ref{sub-sec:method-distill}, and $q_{i}$ is the class prediction of $M_2$ for $d_{i}$ with the same temperature as that of $M_1$. $\lambda$ is a hyper-parameter to govern the relative weight of the two terms ($0 \leq \lambda \leq 1$). Since the gradients of the second term in Equation \ref{eq:loss} scale by $\frac{1}{T^2}$, in order to balance the impact of the two terms in back-propagation, we multiply these gradients by $T^2$--see Equation \ref{eq:softmax}.

The first term in Equation \ref{eq:loss} is the cross entropy between the ground truth labels and the class probabilities of $M_1$. The second term is the cross entropy between the class probabilities of $M_2$ and $M_1$. This objective function is an effort to keep a balance between the information that is transferred from the previous iterations and the information that is extracted from the set of labeled documents~$L$.

% If we take the derivative of $\mathcal{L}$ with respect to the \textit{i-th} input of the softmax layer in $M_1$ when we calculate the gradients of the first term in Equation \ref{eq:loss}, we will have:
% \begin{equation} \label{eq:t1-grad}
% \setlength{\jot}{0pt}
% \setlength{\abovedisplayskip}{0pt}
% \setlength{\belowdisplayskip}{0pt}
% \medmuskip=0mu
% \thinmuskip=0mu
% \thickmuskip=0mu
% \nulldelimiterspace=0pt
% \scriptspace=0pt
% \begin{split}
% \frac{\partial \mathcal{L}_1}{\partial z_i}= &(1-\lambda)(-\sum_{i=1}^{N}[y_{i} \frac{\partial \log a_{i}}{\partial z_i}+(1-y_{i})\frac{\partial \log (1-a_{i})}{\partial z_i}]) \\
% &=(1-\lambda)(-\sum_{i=1}^{N}[y_{i} \frac{1}{a_i}~\frac{\partial a_{i}}{\partial z_i}+(1-y_{i}) \frac{1}{1-a_{i}}~\frac{\partial (1-a_{i})}{\partial z_i}]) \\
% &=(1-\lambda)(-\sum_{i=1}^{N}[y_{i} (1-a_i)-(1-y_{i}) a_i]) \\
% &=(1-\lambda)\sum_{i=1}^{N}[a_i - y_{i}]
% \end{split}
% \end{equation}
% Correspondingly, if we calculate the derivative of the second term in $\mathcal{L}$ with respect to the \textit{i-th} input of the softmax layer in $M_1$ when we have the temperature $T$, as we described in Equation \ref{eq:softmax}, we will get:
% \begin{equation} \label{eq:t2-grad}
% \setlength{\jot}{0pt}
% \setlength{\abovedisplayskip}{0pt}
% \setlength{\belowdisplayskip}{0pt}
% \medmuskip=0mu
% \thinmuskip=0mu
% \thickmuskip=0mu
% \nulldelimiterspace=0pt
% \scriptspace=0pt
% \begin{split}
% \frac{\partial \mathcal{L}_2}{\partial z_i}\approx \lambda \frac{1}{T^2} \sum_{i=1}^{N}[a'_i - q_{i}]
% \end{split}
% \end{equation}

In Section \ref{sec:result} we demonstrate that the ideas proposed in this section greatly mounts the resistance of \METHOD to the noise in the pseudo-labels. These ideas are related to two categories of studies: 1) The studies on pretraining neural networks \cite{pretraining-1, pretraining-2}. 2) The studies on curriculum learning \cite{curriculum}. Researchers \cite{pretraining-1, pretraining-2} in both NLP and the vision community have shown that \textit{pretraining} a neural network with out-of-domain data and then \textit{finetuning} it with the target data can significantly contribute to the performance. These two steps are analogous to the two stages that we described in this section. Additionally, our work is also closely related to the idea of curriculum learning \cite{curriculum}, where it is shown that a learner can leverage the order of the training examples to learn more efficiently. Even though \METHOD employs this mechanism, the criterion to determine the order of the training examples is not based on the properties of the data points but is based on the source of the labels.

\subsection{Right Trapezoidal Learning Rates} \label{sub-sec:method-lr}
In the previous section, we employed an approach to mitigate the semantic drift problem by exploiting the catastrophic forgetting phenomenon. This two-stage strategy creates a suitable opportunity for enhancing the optimization process. Since the pseudo-labels are potentially noisy, we propose to use this set to explore the hypothesis space and detect the region that contains a better local-optima. Thereafter, the set of labeled documents, which are noise-free, can be used to detect the target local-optima.

\begin{figure}
\centering
\includegraphics[width=2in]{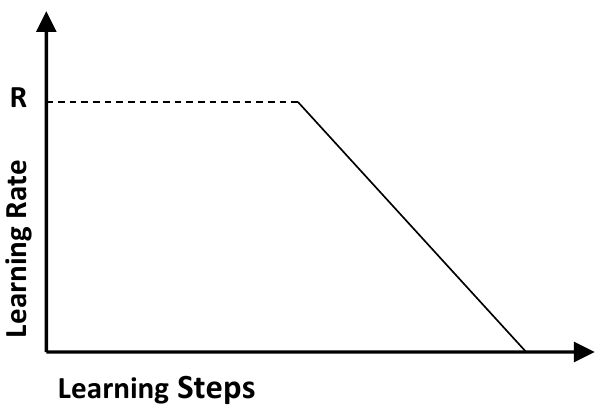}
\caption{{\small The \METHOD learning rate schedule. The dashed horizontal line is the learning rate of the network during the training by the pseudo-labels, and the slanted line is the learning rate during the training by the labeled documents.}} \label{fig:lr}
\end{figure}

Given the argument above, we propose to use a \textit{right trapezoidal} learning rate--illustrated in Figure \ref{fig:lr}--as follows:
\begin{equation} \nonumber
\setlength{\jot}{0pt}
\setlength{\abovedisplayskip}{0pt}
\setlength{\belowdisplayskip}{0pt}
\medmuskip=0mu
\thinmuskip=0mu
\thickmuskip=0mu
\nulldelimiterspace=0pt
\scriptspace=0pt
\begin{split}
\eta_t=\left\{\begin{matrix}
R & batch_t \subset C \\ 
R~-~R~\frac{t-b_C}{b_L} & batch_t \subset L
\end{matrix}\right.
\end{split}
\end{equation}
where $t$ denotes the current time step, and $batch_t$ is the current batch of documents being processed. $\eta_t$ is the current learning rate, $R$ is the initial learning rate, $C$ is the set of pseudo-labels, $L$ is the set of labeled documents, $b_C$ is the number of pseudo-label batches, and $b_L$ is the number of labeled batches.

Our proposed learning rate is composed of two phases: 1) A fixed learning rate--the dashed line in Figure \ref{fig:lr}--where the pseudo-labels are used to train the model $M_1$--see Algorithm \ref{alg:summary}. In this stage, the network parameters can freely update, and therefore, the learner can essentially explore the hypothesis space. 2) A gradually decreasing learning rate--the solid slanted line in Figure \ref{fig:lr}--where the labeled documents are used to further train the network. In this stage, the optimizer settles down, therefore, we use the noise-free labeled documents, since even a small perturbation in the data may cause a significant loss. Having a two-phase learning rate also organically integrates with our two-stage semi-supervised learning procedure. Since the gradual reduction in the learning rate, prevents the objective of the second task from completely erasing the knowledge transferred from the previous iterations.
\subsection{Inertial Class Distributions} \label{sub-sec:method-inertia}
Semi-supervised learning models rely on unlabeled data as their primary source of information. While these methods have obtained promising results, they are inherently prone to overfitting on the irregularities in the unlabeled data. Introducing an inductive bias \cite{ml-book} into the semi-supervised learning algorithms is a common approach to increase their robustness. For instance metric regularization \cite{metric-reg} or temporal ensembling \cite{temporal-ensemble} are a few examples. While these techniques can be integrated into \METHOD\!, in this section, we opt to explore a new direction.

We hypothesize that the class probability distribution of the randomly selected set of unlabeled documents--Algorithm \ref{alg:summary}, Line \ref{alg-line:retrain}--should evolve slowly and avoid abrupt transitions across iterations. This is a harsh assumption, since this probability distribution also depends on the drawn samples. However, we argue that an abrupt change in this distribution can be the sign of an influx of noisy pseudo-labels in the previous iterations. Thus, we aim to prohibit such changes. To achieve this, we assume the distribution of the class probabilities is a random process dynamically evolving across the iterations, and the class probability distribution of the selected unlabeled documents in every iteration--$M_2(C)$ in Algorithm \ref{alg:summary}--is a sample from the underlying random variables.

For simplicity, we assume the process consists of only a family of two Gaussian random variables $S^{+}$ and $S^{-}$, where $S^{+}$ is the state of the positive pseudo-labels, and $S^{-}$ is the state of the negative pseudo-labels. The sample mean and variance of $S^{+}$ in the iteration $t$ (i.e., $S_{t}^{+}$) are given by:
\begin{equation} \nonumber
\setlength{\jot}{0pt}
\setlength{\abovedisplayskip}{0pt}
\setlength{\belowdisplayskip}{0pt}
\medmuskip=0mu
\thinmuskip=0mu
\thickmuskip=0mu
\nulldelimiterspace=0pt
\scriptspace=0pt
\begin{split}
&\mu_{t} = \frac{\sum_{i=1}^{n}p_i^t} {n} \\
&\sigma_{t}^2 = \frac{\sum_{i=1}^{n}(p_i^t - \mu_{t})^2} {n}
\end{split}
\end{equation}
where $n$ is the number of positive pseudo-labels in the iteration $t$, and $p_i^t$ is the probability of the \textit{i-th} positive pseudo-label belonging to the positive class--it is clear that $0.5 \leq p_i^t$, because the sample is positive. Correspondingly, the sample mean and variance of $S^{-}$ in the iteration $t$ (i.e., $S_{t}^{-}$) are given by:
\begin{equation} \nonumber
\setlength{\jot}{0pt}
\setlength{\abovedisplayskip}{0pt}
\setlength{\belowdisplayskip}{0pt}
\medmuskip=0mu
\thinmuskip=0mu
\thickmuskip=0mu
\nulldelimiterspace=0pt
\scriptspace=0pt
\begin{split}
&\gamma_{t} = \frac{\sum_{i=1}^{m}q_i^t} {m} \\
&\varphi_{t}^2 = \frac{\sum_{i=1}^{m}(q_i^t - \gamma_{t})^2} {m}
\end{split}
\end{equation}
where $m$ is the number of negative pseudo-labels in the iteration $t$, and $q_i^t$ is the probability of the \textit{i-th} negative pseudo-label belonging to the negative class--note that $0.5 \leq q_i^t$ and also note that for every pseudo-label $p_i^t+q_i^t=1$.

In the iteration $t+1$, the sample distributions of the random variables $S^+$ and $S^-$ proceed to $S_{t+1}^{+} \sim \mathcal{N}(\mu_{t+1},\,\sigma_{t+1}^2)$ and $S_{t+1}^{-} \sim \mathcal{N}(\gamma_{t+1},\,\varphi_{t+1}^2)$. These updates can be due to the randomness in the model initialization, the randomness in the selected set of unlabeled documents in the iteration $t$, or partially due to the noisy pseudo-labels introduced in the iteration $t$. More specifically, the misclassifications of the model $M_2$ in the iteration $t$--see Algorithm \ref{alg:summary}--which were subsequently used to pretrain the model $M_1$, and ultimately distorted the class distribution of the set of pseudo-labels in the iteration $t+1$. To dampen the impact of this noise, we define two Gaussian distributions $\hat{S}_{t+1}^{+}$ and $\hat{S}_{t+1}^{-}$ as the linear combination of the class distributions in the iterations $t$ and $t+1$, and project\footnote{No projection is performed in the first iteration.} the pseudo-labels in $S_{t+1}^{+}$ into $\hat{S}_{t+1}^{+}$, and the pseudo-labels in $S_{t+1}^{-}$ into $\hat{S}_{t+1}^{-}$. Thus:
\begin{equation} \label{eq:moment}
\setlength{\jot}{0pt}
\setlength{\abovedisplayskip}{0pt}
\setlength{\belowdisplayskip}{0pt}
\medmuskip=0mu
\thinmuskip=0mu
\thickmuskip=0mu
\nulldelimiterspace=0pt
\scriptspace=0pt
\begin{split}
& \hat{S}_{t+1}^{+} = \alpha~S_{t}^{+} + (1 - \alpha)~S_{t+1}^{+} \\
& \hat{S}_{t+1}^{-} = \alpha~S_{t}^{-} + (1 - \alpha)~S_{t+1}^{-}
\end{split}
\end{equation}
where $\alpha$ is a hyper-parameter to govern the rate at which the probability distributions can evolve in every iteration. The new distributions $\hat{S}_{t+1}^{+}$ and $\hat{S}_{t+1}^{-}$ are defined between the class distributions in the iteration $t$ and $t+1$. The hyper-parameter $\alpha$ determines the degree at which the pseudo-labels in the iteration $t+1$ are perturbed to resemble the pseudo-labels in the iteration $t$. By employing this mechanism, the sudden abrupt changes in the distribution of pseudo-labels are avoided. We perform this step after we generate the pseudo-labels using the model $M_2$, and before using this set to pretrain the model $M_1$--Algorithm \ref{alg:summary}, Line \ref{alg-line:retrain}.

In Section \ref{sec:result}, we show that \METHOD algorithm, along the techniques that we introduced in the sections \ref{sub-sec:method-distill}, \ref{sub-sec:method-pretrain}, \ref{sub-sec:method-lr}, and \ref{sub-sec:method-inertia} achieves the state-of-the-art results in multiple settings. In the next section, we describe our datasets, baselines, and training setup.

%% file: doc-experiment-setup.tex
We begin this section by describing the datasets that we used, then we provide a brief overview of the baseline models, and finally review the detail of the experiments.

\subsection{Datasets} \label{sub-sec:datasets}
We evaluate \METHOD on three Twitter text classification tasks\footnote{Please refer to the cited articles for the analysis and discussion on the difficulties of these tasks, we skip this subject.}: 1)~Adverse Drug Reaction monitoring (ADR). In this task, the goal is to detect the tweets that report an adverse drug effect. We used the dataset introduced in \cite{SMM4H-2019} prepared for the ACL 2019 SMM4H Shared Task. 2)~Crisis Report Detection (CRD). In this task, the goal is to detect the tweets that mention an event related to natural disasters. We used the dataset introduced in \cite{crisis-domain-adapt} about the 2015 Nepal earthquake. 3)~Product Consumption Pattern identification (PCP). In this task, the goal is to identify the tweets that report the usage of a product. We used the dataset introduced in \cite{vaccine-dt}, which is about receiving an influenza vaccine.

\begin{table}
\centering
\small
\begin{tabu}{p{0.19in} p{0.30in} p{0.25in} p{0.25in}  p{0.30in} p{0.25in} p{0.25in} }
\cline{1-7} & \multicolumn{3}{c}{\textbf{Training}}  &
 \multicolumn{3}{c}{\textbf{Test}}  \\ 
\cmidrule(l){2-4} \cmidrule(l){5-7}
\multicolumn{1}{c}{\textbf{Dataset}} & 
Tweets & Neg & Pos & Tweets & Neg & Pos \\ \hline
\multicolumn{1}{c}{\ADRDT} & 20624 & 91\% & 9\% & 4992 & 92\% & 8\% \\ 
\multicolumn{1}{c}{\EARTHQUAKEDT} & 8166 & 53\% & 47\% & 3502 & 53\% & 47\% \\ 
\multicolumn{1}{c}{\PRODUCTDT} & 4503 & 69\% & 31\% & 2114 & 78\% & 22\% \\ \hline
\end{tabu}
\caption{{\small Summary of \text{\ADRDT}, \text{\EARTHQUAKEDT}, and \PRODUCTDT datasets.}} \label{tbl:datasets}
\vspace{-0.5cm}
\end{table}
The \ADRDT and \EARTHQUAKEDT datasets are released with pre-specified training and test sets. In \PRODUCTDT dataset we used the tweets published in 2013 and 2014 for the training set, and the tweets published in 2015 and 2016 for the test set. Table \ref{tbl:datasets} summarizes the datasets. We see that \EARTHQUAKEDT dataset is balanced and \ADRDT dataset is highly imbalanced. The \EARTHQUAKEDT dataset is released along a set of unlabeled tweets. For the other two datasets, we used the Twitter API and crawled 10,000 related tweets for each one to be used as the unlabeled sets (the set $U$ in Algorithm \ref{alg:summary}). For \ADRDT dataset we used the drug names to collect the unlabeled set and for \PRODUCTDT dataset we used the query ``flu AND (shot OR vaccine)'' to collect the set.

\subsection{Baselines} \label{sub-sec:baselines}
We compare our model with six baselines.

\noindent\textbf{\textit{Baseline.}} The setting for evaluating semi-supervised learning models should be realistic. Pretrained contextual language models are the primary ingredient of the state-of-the-art text classifiers. Thus, we used BERT \cite{bert} as the naive baseline, and also as the underlying classifier for all the other baselines. Note that this makes any improvement over the base classifier very challenging, since the improvement should be additive. We train this model on the set of labeled documents, and evaluate on the test set. We used the published pretrained \textit{base} variant, followed by one fully connected layer and one softmax layer. We used the Pytorch implementation \cite{bert-impl} of BERT; the settings are identical to the suggestions in \cite{bert}.

\noindent\textbf{\textit{Self-training.}} We included the regular self-training algorithm \cite{self-train}, where in each iteration the top pseudo-labels, subject to a minimum threshold confidence, are selected and added to the labeled set. We used one instance of \textit{Baseline} in this algorithm.

\noindent\textbf{\textit{Tri-training+.}} We included a variant of tri-training algorithm called tri-training with disagreement \cite{tri-train-d}. In \cite{neural-self-train-baselines}, the authors show that this model is a very strong baseline for semi-supervised learning. We used three instances of \textit{Baseline} in this algorithm.

\noindent\textbf{\textit{Mutual-learning.}} We included the model introduced in \cite{mutual-learn}. This model is an ensemble, and is based on the idea that increasing the entropy of the class predictions improves generalization \cite{entropy-inc}. We used two instances of \textit{Baseline} in this model--in the parallel setting.

\noindent\textbf{\textit{Spaced-rep.}} We included the model introduced in \cite{self-train-queue}. This model employs a queuing technique along a validation set to select the unlabeled documents that are easy and also informative for the task. We used our own implementation of this model.

\noindent\textbf{\textit{Co-Decomp.}} We included the framework introduced in \cite{co-decomp}. This model uses domain knowledge to decompose the task into a set of subtasks to be solved in a multi-view setting. We used the keyword level representations and sentence level representations as the two views. We used two instances of \textit{Baseline} in this algorithm.

\noindent\textbf{\textit{\METHOD\!.}} The model that we introduced in Section \ref{sec:method}. We used two instances of \textit{Baseline} as $M_1$ and $M_2$.

\subsection{Experimental Details} \label{sub-sec:exp-detail}
To evaluate the models in the semi-supervised setting, we sampled a small subset of the training sets\footnote{Using the entire set of labeled tweets turns the classification task into a supervised problem, which is not the subject of our study.} and did not use the rest of the tweets. Note that the remaining set was not used as the unlabeled data either--see Section \ref{sub-sec:datasets} for the description of the unlabeled sets. To sample the data, we used a stratified random sampling to preserve the ratio of the positive to the negative documents. We also ensured that the initial labeled set is identical for all the models. We repeated all the experiments 3 times with different random seeds. We will report the average across the runs. All the baseline models use throttling \cite{semiupervised-nlp-book} with confidence thresholding ($\theta=0.9$). We also linearly increased the size of the sample set \cite{sample-linear}, however, did not add more than 10\% of the current training set in each iteration.

In our experiments we observed that the performances of \textit{self-training} and \textit{Co-Decomp} degrade if we use the entire set of unlabeled data--due to the semantic drift problem. Thus, we assumed the number of the iterations in these algorithms is a hyper-parameter and used 20\% of the labeled set as the validation set to find the best value. \textit{Tri-training+} has an internal stopping criterion. \textit{Mutual-learning} uses the unlabeled data as a regularizer. \textit{Spaced-rep} requires a validation set for the stopping criterion and also for the candidate selection. Thus, in this model we used 20\% of the labeled set as the validation set. We also set the number of queues to 6, the rest of the settings are identical to what is used in \cite{self-train-queue}.

Since we are experimenting in the semi-supervised setting, we did not do full hyper-parameter tuning. We used the training set in \PRODUCTDT dataset and searched for the optimal values of $\lambda$ in Equation \ref{eq:loss} and $\alpha$ in Equation \ref{eq:moment}. Their best values are 0.3 and 0.1 respectively. We set the step size $k$ in Algorithm \ref{alg:summary} to 2,000 and the temperature $T$ in Equation \ref{eq:softmax} to 3. In our two-stage training procedure the goal of the first step is the model initialization, thus we trained the network for only 1 epoch. In the rest of the cases, including in our model and the baselines we trained the models for 3 epochs. The only exception is \textit{Space-rep}, which requires a certain number of training epochs with early stopping. To train BERT in all of the cases we used a setting identical to that of the reference \cite{bert}--we set the batch-size to 32. Following the argument in \cite{perf-metric}, we used F1 in the positive class to tune the models. In the next section, we report average F1, Precision, and Recall of the models across the runs.

%% file: doc-results.tex
We begin this section by reporting the main results. Then we present a series of experiments that we carried out to better understand the properties of \METHOD\!.

\vspace{-0.1cm}
\subsection{Main Results} \label{sub-sec:main-results}
\begin{table*}
\centering
\small
\begin{tabu}{p{0.2in} p{0.5in} p{0.3in} p{0.45in} p{0.35in}  p{0.3in} p{0.45in} p{0.35in}  p{0.3in} p{0.45in} p{0.35in} }
 \cline{1-11} & & \multicolumn{3}{c}{\textbf{\ADRDT dataset}}  &
 \multicolumn{3}{c}{\textbf{\EARTHQUAKEDT dataset}} &
 \multicolumn{3}{c}{\textbf{\PRODUCTDT dataset}} \\ 
\cmidrule(l){3-5} \cmidrule(l){6-8}  \cmidrule(l){9-11}
\multicolumn{1}{c}{\textbf{\# Tweets}} & \multicolumn{1}{c}{\textbf{Model}} & 
\textbf{F1} & \textbf{Precision} & \textbf{Recall} &
\textbf{F1} & \textbf{Precision} & \textbf{Recall} &
\textbf{F1} & \textbf{Precision} & \textbf{Recall}
\\ \hline
\multicolumn{1}{c}{\multirow{7}{*}{300}} & \multicolumn{1}{c}{\textit{Baseline}} & 0.238 & 0.237 & 0.342 & 0.715 & 0.692 & 0.749 & 0.728 & 0.696 & 0.770 \\ 
\multicolumn{1}{c}{} & \multicolumn{1}{c}{\textit{Self-training}} & 0.303 & 0.269 & 0.350 & 0.728 & 0.697 & 0.762 & 0.731 & 0.675 & 0.798 \\ 
\multicolumn{1}{c}{} & \multicolumn{1}{c}{\textit{Tri-training+}} & 0.306 & 0.236 & \textbf{0.448} & 0.735 & 0.680 & 0.799 & 0.734 & 0.659 & \textbf{0.828} \\ 
\multicolumn{1}{c}{} & \multicolumn{1}{c}{\textit{Mutual-learning}} & 0.024 & \textbf{0.707} & 0.012 & \textbf{0.743} & 0.685 & \textbf{0.814} & 0.753 & \textbf{0.778} & 0.730 \\ 
\multicolumn{1}{c}{} & \multicolumn{1}{c}{\textit{Spaced-rep}} & 0.258 & 0.248 & 0.277 & 0.721 & 0.650 & 0.811 & 0.727 & 0.701 & 0.760 \\
\multicolumn{1}{c}{} & \multicolumn{1}{c}{\textit{Co-Decomp}} & 0.310 & 0.288 & 0.356 & 0.728 & \textbf{0.722} & 0.735 & 0.754 & 0.756 & 0.758 \\
\multicolumn{1}{c}{} & \multicolumn{1}{c}{\textit{\METHOD}} & \textbf{0.397} & 0.370 & 0.440 & 0.737 & 0.704 & 0.772 & \textbf{0.766} & 0.757 & 0.777 \\ \hline

\multicolumn{1}{c}{\multirow{7}{*}{500}} & \multicolumn{1}{c}{\textit{Baseline}} & 0.312 & 0.253 & 0.411 & 0.746 & 0.735 & 0.760 & 0.740 & 0.704 & 0.782 \\ 
\multicolumn{1}{c}{} & \multicolumn{1}{c}{\textit{Self-training}} & 0.335 & 0.300 & 0.387 & 0.737 & \textbf{0.765} & 0.714 & 0.741 & 0.739 & 0.745 \\ 
\multicolumn{1}{c}{} & \multicolumn{1}{c}{\textit{Tri-training+}} & 0.365 & 0.298 & 0.480 & 0.747 & 0.707 & \textbf{0.793} & 0.758 & 0.697 & \textbf{0.833} \\ 
\multicolumn{1}{c}{} & \multicolumn{1}{c}{\textit{Mutual-learning}} & 0.108 & \textbf{0.638} & 0.059 & 0.751 & 0.730 & 0.773 & 0.767 & \textbf{0.811} & 0.728 \\ 
\multicolumn{1}{c}{} & \multicolumn{1}{c}{\textit{Spaced-rep}} & 0.295 & 0.274 & 0.417 & 0.728 & 0.694 & 0.775 & 0.737 & 0.693 & 0.788 \\
\multicolumn{1}{c}{} & \multicolumn{1}{c}{\textit{Co-Decomp}} & 0.345 & 0.313 & 0.388 & 0.749 & 0.746 & 0.752 & 0.766 & 0.771 & 0.764 \\  
\multicolumn{1}{c}{} & \multicolumn{1}{c}{\textit{\METHOD}} & \textbf{0.420} & 0.376 & \textbf{0.483} & \textbf{0.752} & 0.718 & 0.789 & \textbf{0.787} & 0.784 & 0.792 \\ \hline
\end{tabu}
\caption{F1, precision, and recall of \METHOD in \ADRDT\!, \EARTHQUAKEDT\!, and \PRODUCTDT\! datasets compared to the baseline models. The models were trained on 300 and 500 labeled user postings.} \label{tbl:result-semi}
\vspace{-0.5cm}
\end{table*}
Table \ref{tbl:result-semi} reports the performance of \METHOD in comparison to the baselines under two sampling quantities--i.e, 300 and 500 initial random tweets--in the three datasets. We see that in all of the cases \METHOD is either the top model or on a par with the top model. The difference in \ADRDT dataset is substantial, however, in \EARTHQUAKEDT dataset the difference is very small. \ADRDT is an imbalanced dataset. Our case by case inspections also showed that the positive tweets in this dataset are very diverse, which makes the models very susceptible to the number of training examples. We also see that \textit{Mutual-learning} completely fails in this dataset. Our experiments showed that this is due the the skewed class distributions in this dataset\footnote{We built two imbalanced datasets by subsampling from \EARTHQUAKEDT and \PRODUCTDT datasets, this model also failed in these cases.}. Surprisingly, we see that \textit{Spaced-rep} is performing poorly in the experiments, even though this model was evaluated on social media tasks before \cite{self-train-queue}. We believe the reason is as follows: This model relies on the number of training epochs to construct its internal data structure for ranking the candidate tweets. When the underlying classifier is a pretrained language model, e.g., bert, increasing the number of epochs may result in overfitting and therefore, contradicts the purpose. On the other hand, early stopping also prevents the model from separating the informative from uninformative tweets.

\subsection{Empirical Analysis} \label{sub-sec:analysis}
\begin{figure}
\centering
\includegraphics[width=2.5in]{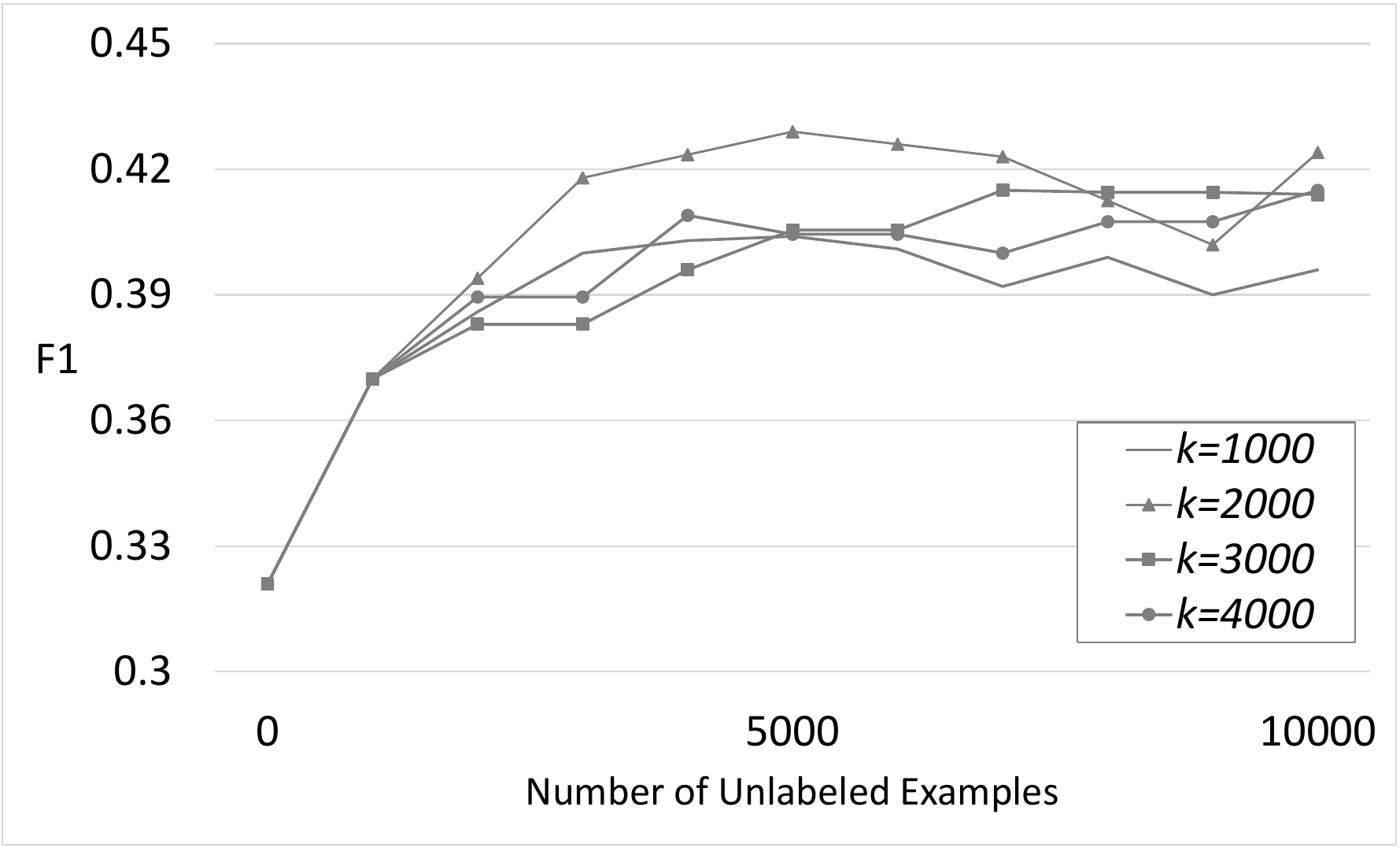}
\caption{{\small F1 of the resulting classifier in every iteration of \METHOD with different values of $k$--the number of randomly selected pseudo-labels. The middle values are interpolated. The results are in the test set of \ADRDT dataset.}} \label{fig:curve-unlabeled}
\end{figure}

We begin this section by reporting the effect of the step size $k$ on \METHOD\!--see Algorithm \ref{alg:summary}. Table \ref{tbl:pseudo-k} reports F1, precision, and recall of \METHOD at varying step sizes in the test set of \ADRDT dataset. Since this datset is the largest one, we report all of the experiments in this dataset. We see that the performance improves up to the step size of 3000 unlabeled tweets per iteration. We still do not have a concrete explanation to justify this trend, since it is natural to expect the smaller step sizes yield better results. One reason may be that if the set of pseudo-labels is small, the network can perfectly learn the noise in the set during the pretraining. In Section \ref{sub-sec:method-pretrain} we argued that the two-stage training can cope with the semantic drift problem. To support this argument, we report the performance of the middle classifiers $M_1$ at the end of every iteration. Figure \ref{fig:curve-unlabeled} reports the performances during the training for varying step sizes. We see that for none of the step sizes the performance drops as the number of unlabeled tweets grows--the typical symptom of semantic drift.
\begin{table}
\centering
\small
\begin{tabu}{p{0.5in}  p{0.45in} p{0.45in} p{0.45in} } \hline
\textbf{\textit{k}} & \textbf{F1} & \textbf{Precision} & \textbf{Recall} \\ \hline
1000 & 0.395 & 0.306 & 0.565 \\ 
2000 & 0.420 & 0.376 & 0.483 \\ 
3000 & 0.428 & 0.386 & 0.485 \\ 
4000 & 0.413 & 0.347 & 0.537 \\ \hline 
\end{tabu}
\caption{{\small
%F1, precision, and recall of
Results of \METHOD with different values of $k$--the number of randomly selected pseudo-labels--in the test set of \ADRDT dataset. The models began with 500 labeled user postings.}} \label{tbl:pseudo-k}
\vspace{-0.5cm}
\end{table}

\METHOD relies on iterative distillation--Section \ref{sub-sec:method-distill}--to transfer knowledge from one iteration to the next one. Model distillation leverages the temperature $T$ in the softmax layer, see Equation \ref{eq:softmax}. It is informative to find the degree at which this hyper-parameter can affect the learning performance. Table \ref{tbl:temperature} reports the model performance at varying values of the hyper-parameter $T$. We see that the performance peaks at $T=5$. In section \ref{sub-sec:method-pretrain} we proposed an objective function and argued that the second term of the function prevents the hard labels of the training set from erasing the information transferred from the previous iteration. To demonstrate the impact of the second term, in Table \ref{tbl:lambda} we report the model performance at varying values of the hyper-parameter $\lambda$--the weight of the second term. We see that the performance almost gradually improves as we increase $\lambda$ and peaks at $\lambda=0.4$. This is primarily due to the improvement in precision.
\begin{table}
\centering
\small
\begin{tabu}{p{0.3in}  p{0.45in} p{0.45in} p{0.45in} } \hline
\textbf{\textit{T}} & \textbf{F1} & \textbf{Precision} & \textbf{Recall} \\ \hline
2 & 0.422 & 0.361 & 0.514 \\ 
3 & 0.420 & 0.376 & 0.483 \\ 
4 & 0.421 & 0.356 & 0.517 \\ 
5 & 0.433 & 0.382 & 0.506 \\ 
6 & 0.422 & 0.370 & 0.491 \\ \hline 
\end{tabu}
\caption{{\small  %F1, precision, and recall 
Results of \METHOD in the test set of \ADRDT dataset at varying values of the temperature ($T$) for iterative distillation.}} \label{tbl:temperature}
\vspace{-0.5cm}
\end{table}
\begin{table}
\centering
\small
\begin{tabu}{p{0.3in}  p{0.45in} p{0.45in} p{0.45in} } \hline
\textbf{\textit{$\lambda$}} & \textbf{F1} & \textbf{Precision} & \textbf{Recall} \\ \hline
0.1 & 0.425 & 0.357 & 0.529 \\ 
0.2 & 0.428 & 0.355 & 0.541 \\ 
0.3 & 0.420 & 0.376 & 0.483 \\ 
0.4 & 0.438 & 0.377 & 0.531 \\ 
0.5 & 0.421 & 0.350 & 0.534 \\ \hline 
\end{tabu}
\caption{{\small %F1, precision, and recall 
Results of \METHOD in the test set of \ADRDT dataset at varying values of the hyper-parameter ($\lambda$) for our two-stage learning--see Equation \ref{eq:loss}.}} \label{tbl:lambda}
\vspace{-0.5cm}
\end{table}

In Section \ref{sub-sec:method-inertia} we proposed to transform the class probability distribution in the iteration $t+1$ into a new distribution that resembles the distribution in the iteration $t$. We argued that this transformation can help to mitigate the semantic drift problem via constraining the degree at which the pseudo-labels can evolve in every iteration, therefore, can potentially limit the negative impact of noisy pseudo-labels. In Table \ref{tbl:moment} we report the model performance at varying values of the hyper-parameter $\alpha$ in Equation \ref{eq:moment}. This hyper-parameter governs the degree of the transformation. We see that the performance noticeably improves as we increase the value of $\alpha$. Finally, we report an ablation study in Table \ref{tbl:ablation}. In the previous experiments we showed that a better performance in \ADRDT dataset is achievable by a dataset specific hyper-parameter tuning. Nonetheless, we still expect that, with the current hyper-parameters in \ADRDT\!, the ablation study can reveal the relative importance of the \METHOD modules in general. In this experiment, we replaced the two-stage training model (Section \ref{sub-sec:method-pretrain}) with the simple data augmentation of the labeled and pseudo-labels. Additionally, we replaced our right trapezoidal learning rate (Section \ref{sub-sec:method-lr}) with the default slanted learning rate \cite{bert}. We replaced our iterative distillation process (Section \ref{sub-sec:method-distill}) with simply using the hard labels in every iteration. Finally, we deactivated our pseudo-label transformation step (Section \ref{sub-sec:method-inertia}). We see that the two-stage training model and the inertial transform have the highest and the lowest contributions.
\begin{table}
\centering
\small
\begin{tabu}{p{0.3in}  p{0.45in} p{0.45in} p{0.45in} } \hline
\textbf{\textit{$\alpha$}} & \textbf{F1} & \textbf{Precision} & \textbf{Recall} \\ \hline
0.1 & 0.420 & 0.376 & 0.483 \\ 
0.2 & 0.422 & 0.353 & 0.530 \\ 
0.3 & 0.413 & 0.345 & 0.518 \\ 
0.4 & 0.424 & 0.355 & 0.532 \\ 
0.5 & 0.429 & 0.363 & 0.527 \\ \hline 
\end{tabu}
\caption{{\small %F1, precision, and recall 
Results of \METHOD in the test set of \ADRDT dataset at varying values of the hyper-parameter ($\alpha$) for the inertial transformation of the pseudo-labels--see Equation \ref{eq:moment}.}} \label{tbl:moment}
\vspace{-0.5cm}
\end{table}
\begin{table}
\centering
\small
\begin{tabu}{p{0.9in}  p{0.45in} p{0.45in} p{0.45in} } \hline
\textbf{Deactivated Step} & \textbf{F1} & \textbf{Precision} & \textbf{Recall} \\ \hline
two-stage learning & 0.339 & 0.373 & 0.333 \\ 
trapezoidal lr & 0.360 & 0.235 & 0.770 \\ 
iterative distillation & 0.389 & 0.320 & 0.495 \\ 
inertial transform & 0.420 & 0.365 & 0.497 \\ \hline 
\end{tabu}
\caption{{\small %F1, precision, and recall 
Results of \METHOD in the test set of \ADRDT dataset after deactivating the distillation (Section \ref{sub-sec:method-distill}), the two-stage learning (Section \ref{sub-sec:method-pretrain}), the trapezoidal learning rate (Section \ref{sub-sec:method-lr}), and the inertial transformation (Section \ref{sub-sec:method-inertia}).}} \label{tbl:ablation}
\vspace{-0.5cm}
\end{table}

In summary, we showed that \METHOD is the state-of-the-art in multiple settings. The authors in \cite{self-train-queue} show that semi-supervised models--although under domain shift--typically fail when they are evaluated on a different task from what they are initially proposed for. Thus, they conclude that these models should be evaluated in at least two datasets. In this study we evaluated \METHOD in three Twitter datasets. We selected strong baselines, i.e., Tri-training with disagreement \cite{tri-train-d}, Mutual Learning \cite{mutual-learn}, Spaced Repetition \cite{self-train-queue}, and Co-Decomp \cite{co-decomp}, and showed that some of them fail under certain cases. As opposed to these models, we demonstrated that \METHOD is either the best model or on a par with the best model in every setting. We also reported an extensive set of experiments that we carried out to reveal the qualities of \METHOD\!. These experiments empirically supported the claims that we made throughout. 

Our study is not flawless. To avoid imposing any constrain on the underlying classifier, we proposed to randomly draw the unlabeled documents--Algorithm \ref{alg:summary}, Line \ref{alg-line:sample}. However, if one can guarantee certain classifier properties, then perhaps a sophisticated selection policy will be more effective. The application of our framework in other modalities, e.g., image classification, is also an unexplored topic. Future work may investigate these directions.

%% file: doc-conclusion.tex
In this study, we proposed a semi-supervised learning model called \METHOD\!. Our model is inspired by the traditional self-training algorithm. \METHOD employs the properties of neural networks to cope with the inherent problems of self-training. Particularly, it employs an iterative distillation procedure to transfer information across the iterations. It also utilizes a two-stage training model to mitigate the semantic drift problem. Additionally, \METHOD uses an efficient learning rate schedule and a pseudo-label transformation heuristic. We evaluated our model in three publicly available Twitter datasets, and compared with six baselines, including pretrained BERT. The experiments show that our model consistently outperforms the existing baselines.

%% file: doc-ack.tex
\begin{acks}
We thank the anonymous reviewers for their insightful feedback.
\end{acks}